# COVID-19 Monitoring System using Social Distancing and Face Mask Detection on Surveillance video datasets

Sahana Srinivasan[1], Rujula Singh R[2], Ruchita R Biradar[3], Revathi SA[4]
Department of Computer Science and Engineering
R.V. College of Engineering
Bangalore,India

Email: [1]sahanas.cs17@rvce.edu.in, [2]rujulasinghr.cs17@rvce.edu.in, [3]ruchitarbiradar.cs17@rvce.edu.in, [4]revathisa@rvce.edu.in

*Abstract—In the current times, the fear and danger of COVID-19 virus still stands large. Manual monitoring of social distancing norms is impractical with a large population moving about and with insufficient task force and resources to administer them. There is a need for a lightweight, robust and 24X7 video-monitoring system that automates this process. This paper proposes a comprehensive and effective solution to perform person detection, social distancing violation detection, face detection and face mask classification using object detection, clustering and Convolution Neural Network (CNN) based binary classifier. For this, YOLOv3, Density-based spatial clustering of applications with noise (DBSCAN), Dual Shot Face Detector (DSFD) and MobileNetV2 based binary classifier have been employed on surveillance video datasets. This paper also provides a comparative study of different face detection and face mask classification models. Finally, a video dataset labelling method is proposed along with the labelled video dataset to compensate for the lack of dataset in the community and is used for evaluation of the system. The system performance is evaluated in terms of accuracy, F1 score as well as the prediction time, which has to be low for practical applicability. The system performs with an accuracy of 91.2% and F1 score of 90.79% on the labelled video dataset and has an average prediction time of 7.12 seconds for 78 frames of a video.*

## I. INTRODUCTION

Coronaviruses (CoV) are a wide group of viruses which cause illness that range from colds to deadly infections like Middle East Respiratory Syndrome (MERS) and Severe Acute Respiratory Syndrome (SARS) [1]. Globally, as of January 2021, there have been 90,054,813 confirmed cases of COVID-19, including 1,945,610 deaths, reported to WHO [2]. The COVID-19 virus can spread by direct contact(infected people) or indirect contact (contaminated environment). This occurs with respiratory droplets which are droplet particles>5-10 μm in diameter. The droplet transmission usually occurs when a person is in close quarters (within 1 m) with someone who has respiratory symptoms and is therefore at risk of having his/her mouth, nose or eyes exposed to potentially infective respiratory droplets[3]. With the number of infected cases and deaths increasing, it has become crucial to control the spread of the virus as much as possible.

Social distancing should be practiced in combination with other everyday preventive actions to reduce the spread of COVID-19, including wearing masks, avoiding touching your face with unwashed hands, and frequently washing your hands with soap and water for at least 20 seconds. Social distancing is a practice of keeping a safe distance between the person and his/her surrounding[4]. A distance of two meters (six feet) is the proposed distance to ensure the virus doesn't spread by contact [5,6] Following the social distancing norms reduced the contact by 95% for adults of age greater than 60 and 85% for children of age less than 20 [7]. This shows the potential of following the right social distancing norms on "flattening the curve".

COVID-19 spreads mainly from person to person through respiratory droplets. Respiratory droplets travel into the air due to coughing, sneezing, talking, shouting, or singing. These droplets can then land in the mouths or noses of people or they may breathe these droplets in. Thus, use of masks is essential to curb the spread of the virus. Masks are a simple barrier to help prevent the respiratory droplets from reaching others. Studies show that masks reduce the spray of droplets when worn over the nose and mouth.[8,9]

Monitoring the social distancing norms and checking face masks on people manually is not only restrictive with limited resources, but can also lead to human errors. There is an immediate requirement for a solution to administer the virus spread by learning the ideal social distancing norms to be followed by the public. This includes social distance violation detection and face mask classification to determine the safety of the citizens by checking if enough distance is maintained and if face masks are used. This system has a wide range of applicability in various public places with cameras such as in Supermarkets, Petrol Bunks and Traffic signals. This provides processing techniques to leverage a surveillance system for many other applications.

This paper proposes a proactive and lightweight COVID-19 prevention system that uses video surveillance to detect and help authorities ensure that everyone follows social distancing norms and safety norms to reduce spread of virus. The sections in the paper are organised as follows: section II provides the recent work in this field, section III describes the dataset used and provides dataset augmentation and video labelling techniques that can be implemented to increase the size of the dataset and measure the performance of the system, section IV discusses the preprocessing techniques, section V provides the methodology and discusses the various models for person detection, social distancing violation detection, face detection and face mask classification. Section VI provides the results of analysis of the system and section VII gives the future scope of the system. Finally, section VIII gives the conclusion of the paper.

## II. LITERATURE REVIEW

The model proposed by Mohamed Loey et al. [10] is an integration between deep transfer learning (ResNet-50) and classic machine learning algorithms. The last layer in ResNet-50 was removed and replaced with three traditional machine learning classifiers (Support vector machine (SVM), decision tree, and ensemble) to improve their model performance. Among the four types of datasets they used, one dataset contained the largest number of images among the datasets, consisting of real face masks and fake face masks, and consumed more time compared to the others during the training process. There is also no reported accuracy according to related works for this type of dataset. On the training over dataset having real face masks, the decision trees classifier wasn't able to achieve a good classification accuracy (68%) on fake face masks.

A detection network with a backbone, neck and heads is implemented consisting of Resnet as the Backbone, FPN (feature pyramid network) as Neck and classifiers, predictors, estimators, etc [11]. However, due to the limited size of the face mask dataset, it is difficult for learning algorithms to learn better features. There is limited research focusing on face mask detection, and better detection accuracy has to be achieved.

Another approach is a self-developed model named SocialdistancingNet-19 by Rinkal Keniya [12] for detecting the frame of a person and displaying labels for deciding if they are safe or unsafe if the distance is less than a certain value. If a webcam is to be used, it is necessary to have people moving continuously else the detection goes incorrect. This may happen due to the detection method used by the network where the entire frame is detected, and the distance is calculated between the people using centroids (brute force approach).

Shashi Yadav proposed a deep learning approach with Single Shot object Detection (SSD) using MobileNet V2 and OpenCV for social distancing and mask detection [13]. Challenges faced using this approach was that it categorizes people with hand over their faces or occluded with objects as masked. These scenarios are not suited for this model. Here, although an SSD is capable of detecting multiple objects in a frame, it is limited to the detection of a single person in this system.

Most of the papers have tackled either the issue of social distancing monitoring, or face mask detection. And where both were implemented, there is still scope for using better models to achieve better accuracy. Our paper mentions the importance of prediction time, a feature missing in other papers, with prediction time as an evaluation measure, which is necessary for practical applicability of the system . For person detection, the model proposed by the paper is YOLOv3, a state-of-the-art object detection model, followed by DBSCAN to calculate the distances between people and perform clustering to identify if they are far apart or not, which is effectively better than other clustering methods like or brute force distance calculations or k-means, which requires that number of clusters be decided before performing clustering.

For face detection we have DSFD which is a powerful feature extractor with a good accuracy for detecting faces. And for face mask classification, MobileNetV2 was used as it performed the best, compared to Xception and ResNet50. Finally, a mask dataset was created using data augmentation techniques and a labelled video dataset was also created for testing the system by labelling the frames of the video.

## III. DATASET

The dataset collected from existing sources consisting of unmasked and masked faces proved to be insufficient, hence new and two data augmentation techniques were employed to add masks on unmasked faces and to add blurred images. Data augmentation for unmasked faces was performed on datasets with four types of masks as shown in Fig.1.

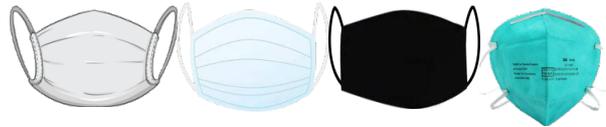

Fig. 1. Masks used for augmentation

The algorithm used for this is depicted in Fig 2. The algorithm begins with identifying the defining points of the outline of a face. The top part of the mask is obtained by locating the nose bridge, and the bottom, left and right parts are identified by locating the chin points of the face. The left and right halves of the mask are prepared with the appropriate size. Using the orientation of the face, the angle of rotation for the mask is calculated. The coordinates for superimposing the mask on the face is calculated and the mask is then placed on the face.

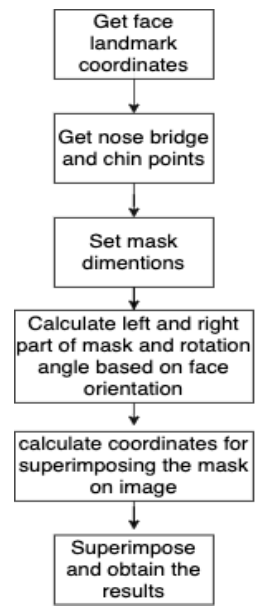

Fig. 2. Steps for dataset generation

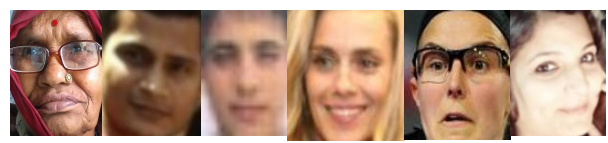

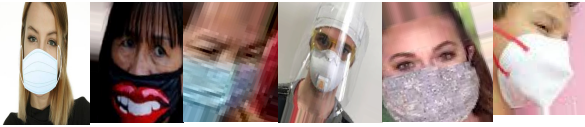

Fig. 3. Unmasked and masked datasets

Due to the nature of the problem, the images are recorded from a surveillance camera which means the faces would be very blurred or occluded. The dataset in contrast had images with clear faces. In Fig 3., to help the model to adapt to the surveillance quality faces, blurring filters such as motion blur, gaussian blur and average blur are used for the second data augmentation.

To perform the blur augmentation, a blurring option is randomly picked among gaussian blur with kernel size ranging from 6 to 10, average blur with kernel size ranging from 3 to 9, motion blur with kernel size ranging from 3 to 10 and no blur option. Gaussian blur[14] is the result of blurring an image with the gaussian function. The pixels nearest the center of the kernel are given more weight than those far away from the center. Motion blur is the apparent streaking of moving objects in a photograph or a sequence of frames by selecting a random direction for the motion blur from vertical, horizontal, main diagonal and anti-diagonal, then convolving the image with the kernel accordingly. Average blur is the result of blurring an image by convolving it with a box filter (normalized). In this process, the central element of the image is replaced by the average of all the pixels in the kernel area.

The final dataset consists of 11,792 labelled images with the classes 0 for no mask and 1 for mask and a 10% of the data was used for testing. Due to the problem being a recent one, there were no video datasets available. There was also no method to measure the performance of the system as a whole on the video. Hence, 30 videos with a duration of 9-15 seconds were procured and individually labelled. This was used to give an unbiased estimation of the performance of the system. The labels chosen were the number of people, the number of people not following social distancing, number of faces detected, and the number of people wearing masks.

## IV. METHODOLOGY

This section discusses the different models used for the system that are listed in Table I which use a wide range of object detection and image classification techniques. Fig. 4 depicts the flow of these models to implement the system. Images are sampled from a surveillance video. The video was processed using a sampling of one frame for every five frames. Since there were about 80 frames per second on average of the videos, some frames could be skipped as the movement of people would not be too drastic within a fraction of a second. Hence, this provides a method to increase the computation speed while not compromising on the performance of the model.

TABLE I. MODELS CONSIDERED FOR INDIVIDUAL COMPONENTS

| Application | Deep learning models considered | Description |
|---|---|---|
| Person detection | YOLOv3-416 | State of the art object detection model for a 416x416 image |
| Social distancing | DBSCAN | Clustering technique |
| Face detection | DSFD RetinaNetMobileNetV1 | State of the art face detection models. |
| Mask classification | ResNet50 Xception MobileNetV2 | CNN based binary classifier |

All images are then resized to 416x416 for the YOLOv3 person detection model and bounding boxes are drawn around them. The centroids of these bounding boxes are determined and social distancing norms are checked using outlier detection with the DBSCAN algorithm with distance parameter set for two meters. Resnet CNN layers are used to extract features from images for face detection[10,15] and the images are resized to 128x128 for the face detection with DSFD and mask classification with MobileNetv2. The image frames are then accumulated to generate the output video.

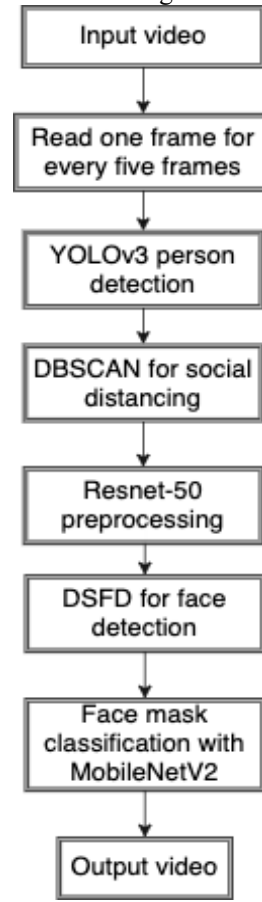

Fig. 4. System flow diagram

### A. PERSON DETECTION

YOLOv3 model was used for person detection [16]. It consists of 53 layers of Darknet-53 trained on Imagenet that acts as a powerful feature extractor and an additional 53 layers for detection giving a total of 106 layered fully convolutional neural network. Fig 5. depicts the YOLOv3 architecture. Anchor box with 3 scales: 13x13, 26x26 and

52x52 are used. These three anchor boxes are used to predict the presence of a person as shown in figure. The output of this model after prediction is a list of bounding boxes along with the confidence of the person class detected.

Non-maximum suppression (NMS) is used to solve the issue of overlapping bounding boxes leading to multiple detections for the same object. The final bounding boxes were selected based on confidence value and NMS threshold whose values were 0.5 and 0.3 respectively. This means only classes with more than 50% confidence are included and all those bounding boxes that have more than 30% overlap with another bounding box are discarded.

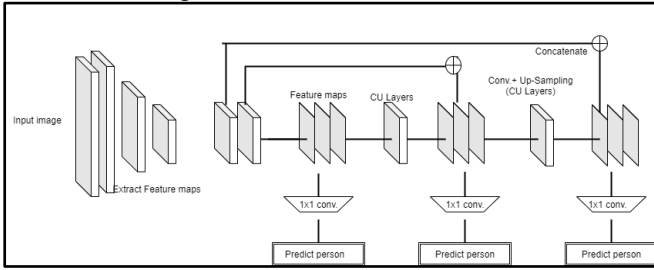

Fig. 5. YOLOv3 architecture

### B. SOCIAL DISTANCING

DBSCAN algorithm was used to check if social distancing is maintained between the persons detected. It is an unsupervised learning algorithm which groups similar points together. DBSCAN, unlike the k-means algorithm, does not require the number of clusters to be set prior training. It also ignores the noisy or outlier points while forming the clusters.

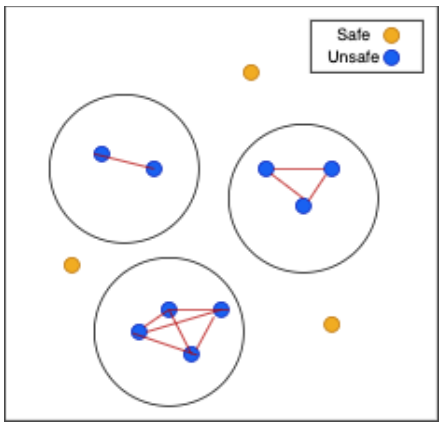

Fig. 6. DBSCAN for social distancing

Fig 6. shows the working of DBSCAN. It is used to perform clustering on the midpoints of the bounding boxes drawn around the detected people. Since social distancing is checked between a minimum of 2 people, minimum required points in the cluster is set to 2, and the distance parameter was set to 200. Taking into consideration person by person, if the distance between them is less than the distance parameter, then they are grouped into a cluster. If a person does not belong to any cluster, then they are categorised as safe and bound with orange boxes. People belonging to a cluster are denoted by red lines between each of them, who are deemed too close, and bound with blue boxes.

### C. FACE DETECTION

For face detection, two pretrained models: DSFD and RetinaNetMobileNetV1 in terms of accuracy and prediction time. RetinaNetMobileNetV1 is a lightweight single shot face detector originally developed for mobile deployment of the face detection models.

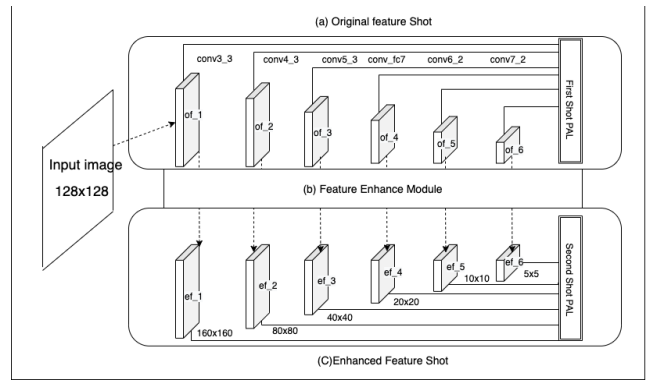

Fig. 7. DSFD architecture

Fig 7. shows DSFD architecture which consists of three components: the first shot detector that consists of convolution layers, the feature enhance module that generates additional features and the second shot detector that incorporates these enhanced features and loss from the first shot detector to give the final predictions. Since a second shot is used, the model performs much better than a single shot detector but is quite slow in prediction.

TABLE II. COMPARISON OF FACE DETECTION MODELS USED

| Models | Accuracy (%) | Detection time on CPU for one second of video | | Detection time on GPU for one second of video | |
|---|---|---|---|---|---|
| | | Seconds | Frames per seconds | Seconds | Frames per seconds |
| RetinaNet MobileNetv1 | 76.7 | 0.65 | 90 | 0.17 | 458 |
| DSFD Detector | 91.1 | 1.92 | 41 | 0.39 | 200 |

There is a tradeoff between accuracy and prediction time as shown in Table II. This is due to the fact that complex models are more accurate but involve a lot of computation and hence take more prediction time. In terms of accuracy, DSFD performs better but has a much higher detection time than RetinaNetMobileNetv1 in Fig 8. Since the images from the video frames were captured from the surveillance camera, the faces are bound to be unclear and blurred. Hence, a compromise on accuracy cannot be made and the DSFD model was chosen.

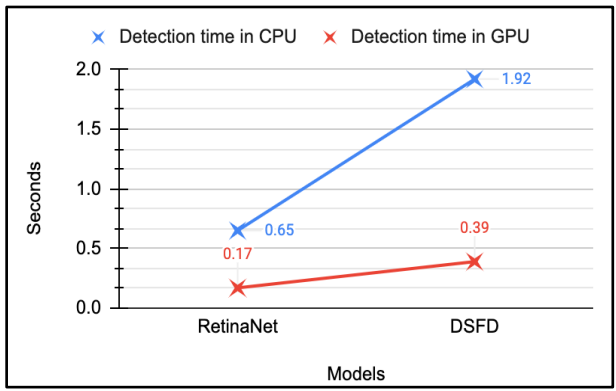

Fig. 8. Detection time for face detection models

## D. *Face mask classification.*

Face mask classification is implemented using CNN binary image classification architecture to check the presence of a mask on the faces detected. A number of models were developed using CNN for mask classification on 128x128 images and their performance in terms of accuracy, precision, recall and F1 scores were compared for class 0 (no mask) and class 1(masked) as shown in Table III. MobileNetV2 was chosen due to its performance in prediction time as well as accuracy as depicted in Fig 9.

TABLE III. COMPARISON OF FACE MASK CLASSIFICATION MODELS

| Models | Accuracy (%) | Precision (%) | Recall (%) | F1-score (%) |
|--------|--------------|---------------|------------|--------------|
| Resnet50 | 47.7 | 49, 46 | 48, 57 | 49, 47 |
| Xception | 50.8 | 51, 51 | 46, 56 | 48, 53 |
| MobilenetV2 | 93.2 | 94.6, 93.80 | 95.7, 94.1 | 95.1, 93.9 |

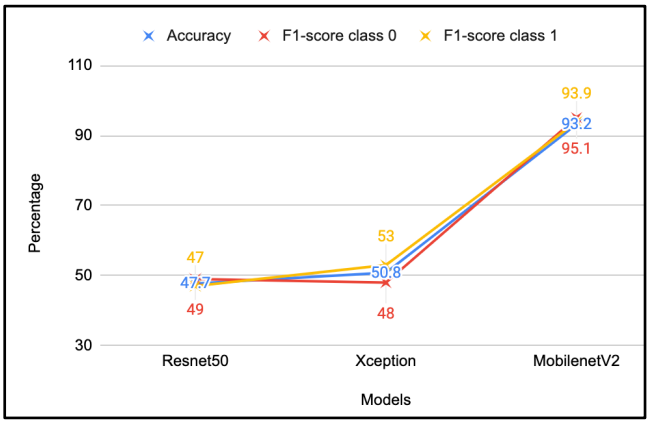

Fig. 9. Performance of face mask classification models

MobileNetV2 provides the best set of performance values with an accuracy of 93.2% on test dataset and 95.6% on training dataset. It has two types of blocks: Residual block with stride of 1 and stride of 2 for downsizing [17]. There are 3 layers for both types of blocks. The first layer is 1×1 convolution with ReLU6. The second layer is the depthwise convolution. The third layer is another 1×1 convolution but without any non-linearity. The architecture is depicted in Fig 10. where C is convolution layer and D is depthwise convolution layer. The output of MobileNetV2 is flattened and passed to a 256 unit fully connected layer with a dropout regularization of 40% followed by another 64 unit fully connected layer and a single output for binary classification.

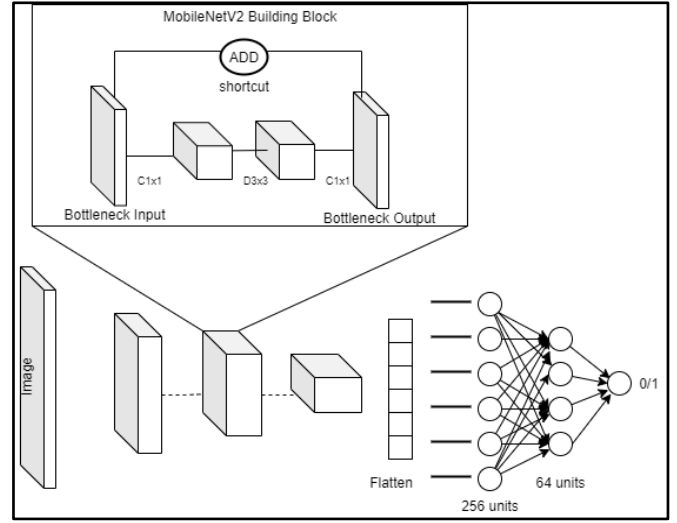

Fig. 10. MobileNetV2 model architecture

The learning curves for the model are shown in Fig 11. The training and validation accuracy increase and the train accuracy is always more than the validation accuracy. Thus, the model is not overfitting. The training and validation loss decrease to a point of stability with a small gap between the training and validation set, the low loss value showing the good fit of the model.

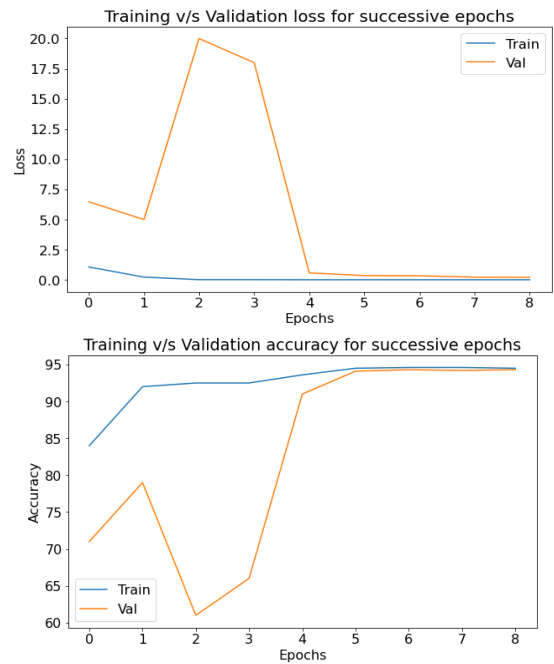

Fig. 11. Learning curves for loss and accuracy

## V. RESULTS

The accuracy and F1 score was calculated using equation 1 and 2. Here TP is the number of true positives, TN is the number of true negatives, FP is the number of false positives and FN is the number of false negatives.

$$\text{Accuracy} = \frac{TP + TN}{(TP + FP) + (TN + FN)} \qquad (1)$$

$$\text{F1 Score} = 2 * \frac{Precision * Recall}{(Precision + Recall)} \qquad (2)$$

The resultant image frame after applying the models on a video frame is shown in Fig 12.

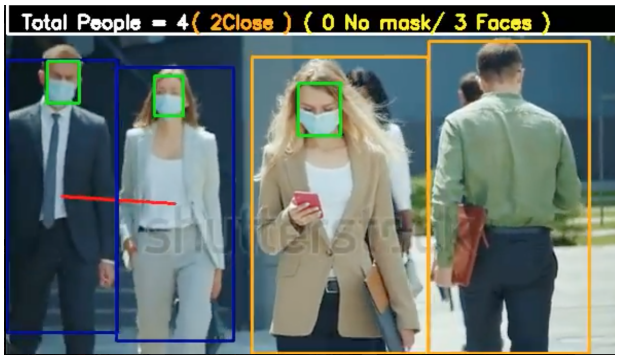

Fig. 12. Snapshot of the generated video frame

Performance was checked on manually labeled videos in Table IV. The F1 score, accuracy and prediction time are depicted to summarize the model performance. The average accuracy of the system is 91.2% and average F1 score is 90.79%. The average prediction time for the overall system is 7.12 seconds for a one second video frame out of which person detection took the most time i.e. 5.24 seconds.

TABLE IV.    SYSTEM PERFORMANCE

| Model | Performance | | |
|---|---|---|---|
| | Accuracy in percentage | F1 Score in percentage | Average time to process 1 second of video (in seconds) |
| Person detection | 93.46 | 93.57 | 5.24 |
| Social distancing | 89.79 | 88.51 | 0.0042 |
| Face detection | 91.01 | 89.05 | 1.92 |
| Mask classification | 90.55 | 88.71 | 0.01 |

## VI. FUTURE SCOPE

Although the system performance is good in terms of accuracy as well as prediction time, following improvement areas are identified: First, the person detection module takes up most of the time in video processing. A simpler person detection algorithm could be developed that takes lesser prediction time with an accuracy on par with the current model. Second, the social distancing calculation and mask classification runs independently and hence parallelism can be used to execute them concurrently. Third, there is a dearth of datasets to be used for such a system and it is not diverse to work for all the situations. For instance, the system sometimes confuses beard with masks due to not having enough negative examples with beards in it. When such datasets become available, a more powerful model can be trained.

## VII. CONCLUSION

This paper provides an efficient solution to monitor social distancing practices in public areas where it is very difficult to monitor manually. Four different modules have been developed for person detection, social distancing recognition, face detection and face mask classification. The system performs reasonably well with an accuracy of 91.2% and average F1 score of 90.79% on the labelled video dataset with average prediction time of 7.12 seconds on a 1 second video (50~90 image frames), where 5.24 seconds is spent on person detection. It also provides data augmentation techniques to deal with the lack of dataset in the community.

***Apology*—** *I, Rujula Singh R, would like to apologize to the research community for the confusion caused by the inconsistency in author lists between multiple versions of this paper. I take full responsibility for this error and will be more diligent in the future to ensure the accuracy and consistency of our research publications. If you would like to cite this paper, please use the conference link provided [18].*

## ACKNOWLEDGEMENT

I, Rujula Singh R, would like to thank Nikhil Nayak for his insights that enhanced my contributions to this paper.

## REFERENCES

[1]    US Centers for Disease Control and Prevention. "Interim pre-pandemic planning guidance: community strategy for pandemic influenza mitigation in the United States: early, targeted, layered use of nonpharmaceutical interventions    Atlanta:    The    Centers;    2007." https://stacks.cdc.gov/view/cdc/11425 [Online; accessed 13 Jan 2021]

[2]    World Health Organization (WHO) "WHO Coronavirus Disease (COVID-19) Dashboard" https://covid19.who.int/, [Online; accessed 13 Jan 2021].

[3]    World Health Organization (WHO) "Modes of transmission of virus causing COVID-19: implications for IPC precaution recommendations "https://www.who.int/news-room/commentaries/detail/modes-of-transmission-of-virus-causing-covid-19-implications-for-ipc-precaution-recommendations [Online; accessed 13 Jan 2021]

[4]    Center for disease control and prevention (CDC) "COVID-19 Social distancing"https://www.cdc.gov/coronavirus/2019-ncov/prevent-getting-sick/social-distancing.html [Online; accessed 13 Jan 2021]

[5]    Manasee Mishra, Piyusha Majumdar; Social Distancing During COVID-19: Will it Change the Indian Society?(2020)

[6]    Marco Cristan, Alessio Del Bue, Vittorio Murino, Fraccesco Setti And Alessandro Vinciarelli The Visual Social Distancing Problem , 2020

[7]    Matrajt L, Leung T. Evaluating the efficacy of social distancing strategies to postpone or flatten the curve of coronavirus disease. Emerg Infect Dis, man (2020)

[8]    World Health Organisation(WHO) "Coronavirus disease (COVID-19) advice for the public: When and how to use masks" https://www.who.int/emergencies/diseases/novel-coronavirus-2019/advice-for-public/when-and-how-to-use-masks [Online; accessed 13 Jan 2021]

[9]    Center for disease control and prevention (CDC) "Considerations for Wearing Masks" https://www.cdc.gov/coronavirus/2019-ncov/prevent-getting-sick/cloth-face-cover-guidance.html [Online; accessed 13 Jan 2021]

[10]    Mohamed Loey , Gunasekaran Manogaran, Mohamed Hamed N. Taha , Nour Eldeen M. Khalifa;A hybrid deep transfer learning model with machine learning methods for face mask detection in the era of the COVID-19 pandemic (2020)

[11]    Mingjie Jiang* . Xinqi Fan* .Hong Yan . RETINAFACEMASK: A FACE MASK DETECTOR(2020)

[12]    Shashi Yadav,Goel Institute of Technology and Management, Dr. A.P.J. Abdul Kalam Technical University ,Deep Learning based Safe Social

Distancing and Face Mask Detection in Public Areas for COVID-19 Safety Guidelines Adherence (2020)

[13]    Rinkal Keniya · Ninad Mehendale, Real-time social distancing detector using Socialdistancing-Net19 deep learning network(2020)

[14]    Indhu Jain, Mr. Sudhir Goswami; A Comparative Study of Various Image Restoration techniques with different types of blur, International Journal Of Research In Computer Applications And Robotics (2015).

[15]    Kaiming He, Xiangyu Zhang, Shaoqing Ren, Jian Sun; Deep Residual Learning for Image Recognition (2015).

[16]    Joseph Redmon, Ali Farhadi, University of Washington ;YOLOv3: An Incremental Improvement.(2018)

[17]    "MobileNetV2: Inverted Residuals and Linear Bottlenecks", The IEEE Conference on Computer Vision and Pattern Recognition (CVPR), 2018

[18]    S. Srinivasan, R. Rujula Singh, R. R. Biradar and S. Revathi, "COVID-19 Monitoring System using Social Distancing and Face Mask Detection on Surveillance video datasets," 2021 International Conference on Emerging Smart Computing and Informatics (ESCI), 2021, pp. 449-455, doi: 10.1109/ESCI50559.2021.9396783.